\documentclass[sigconf]{aamas} 
\usepackage{balance} % for balancing columns on the final page
\usepackage{enumitem,kantlipsum}
\usepackage{color}
\usepackage{subcaption}
\setcopyright{ifaamas}
\acmConference[AAMAS '23]{Proc.\@ of the 22nd International Conference
on Autonomous Agents and Multiagent Systems (AAMAS 2023)}{May 29 -- June 2, 2023}
{London, United Kingdom}{A.~Ricci, W.~Yeoh, N.~Agmon, B.~An (eds.)}
\copyrightyear{2023}
\acmYear{2023}
\acmDOI{}
\acmPrice{}
\acmISBN{}

\title[The Visual Assistant]{Learning a Visually Grounded Memory Assistant}  

\author{
Meera Hahn$^{1,2}$ \quad
Kevin Carlberg$^2$ \quad
Ruta Desai$^2$ \quad
James Hillis$^2$ \\
[0.1in]
$^1$Georgia Institute of Technology \quad
$^2$Facebook Reality Labs \\
[0.1in]
{\normalsize meerahahn@gatech.edu, \{carlberg, rutadesai, jmchillis\}@fb.com}}

\begin{abstract}
We introduce a novel interface for large scale collection of human memory and assistance. Using the 3D Matterport simulator we create a realistic indoor environments in which we have people perform specific embodied memory tasks that mimic household daily activities. This interface was then deployed on Amazon Mechanical Turk allowing us to test and record human memory, navigation and needs for assistance at a large scale that was previously impossible. Using the interface we collect the `The Visually Grounded Memory Assistant Dataset' which is aimed at developing our understanding of (1) the information people encode during navigation of 3D environments and (2) conditions under which people ask for memory assistance. Additionally we experiment with with predicting when people will ask for assistance using models trained on hand-selected visual and semantic features. This provides an opportunity to build stronger ties between the machine-learning and cognitive-science communities through learned models of human perception, memory, and cognition.
\end{abstract}


\newcommand{\BibTeX}{\rm B\kern-.05em{\sc i\kern-.025em b}\kern-.08em\TeX}

\begin{document}

%%% The following commands remove the headers in your paper. For final 
%%% papers, these will be inserted during the pagination process.

\pagestyle{fancy}
\fancyhead{}

%%% The next command prints the information defined in the preamble.

\maketitle 

%%%%%%%%%%%%%%%%%%%%%%%%%%%%%%%%%%%%%%%%%%%%%%%%%%%%%%%%%%%%%%%%%%%%%%%%
\section{Introduction}
\label{sec:intro}
Automated interaction with humans in everyday activity remains a significant challenge for artificial intelligence (AI). Current interactive systems take many forms and operate on different time scales; examples include shopping recommendations, conversational AI, autonomous vehicles, and social robots. These models typically only work within a narrow range of conditions. For example, while autonomous vehicles can interact effectively with other drivers in highway conditions, they are not yet ready for city streets where environments are less predictable. An even more significant challenge is presented by the prospect of all-day wearable augmented reality (AR) glasses: the ideal is an automated system that that can offer assistance in any context. Unlike existing mobile devices, AR wearable glasses could have access to information from the internet \textit{and} detailed information about local physical context, including user actions from a first person viewpoint. Simultaneous access to both sources of information opens new opportunities and challenges for the development of collaborative human--AI systems. The AI required for such a system would likely require “theory of mind” similar to that of humans, whereby people infer goals and cognitive states of others and take strategic, context-dependent actions that may be cooperative or adversarial in nature. 

Here we present a dataset aimed at providing insight into the conditions where people request assistance to recall facts about the local environment. We focus on memory because enhancing human memory is one of the primary uses of computing technology. Our data provides insight into (1) the kind of features people encode during navigation, (2) the difficulty of different types of questions, and (3) conditions under which people will ask for assistance. 

\begin{figure}[htp]
\centering
\includegraphics[width=8.5cm]{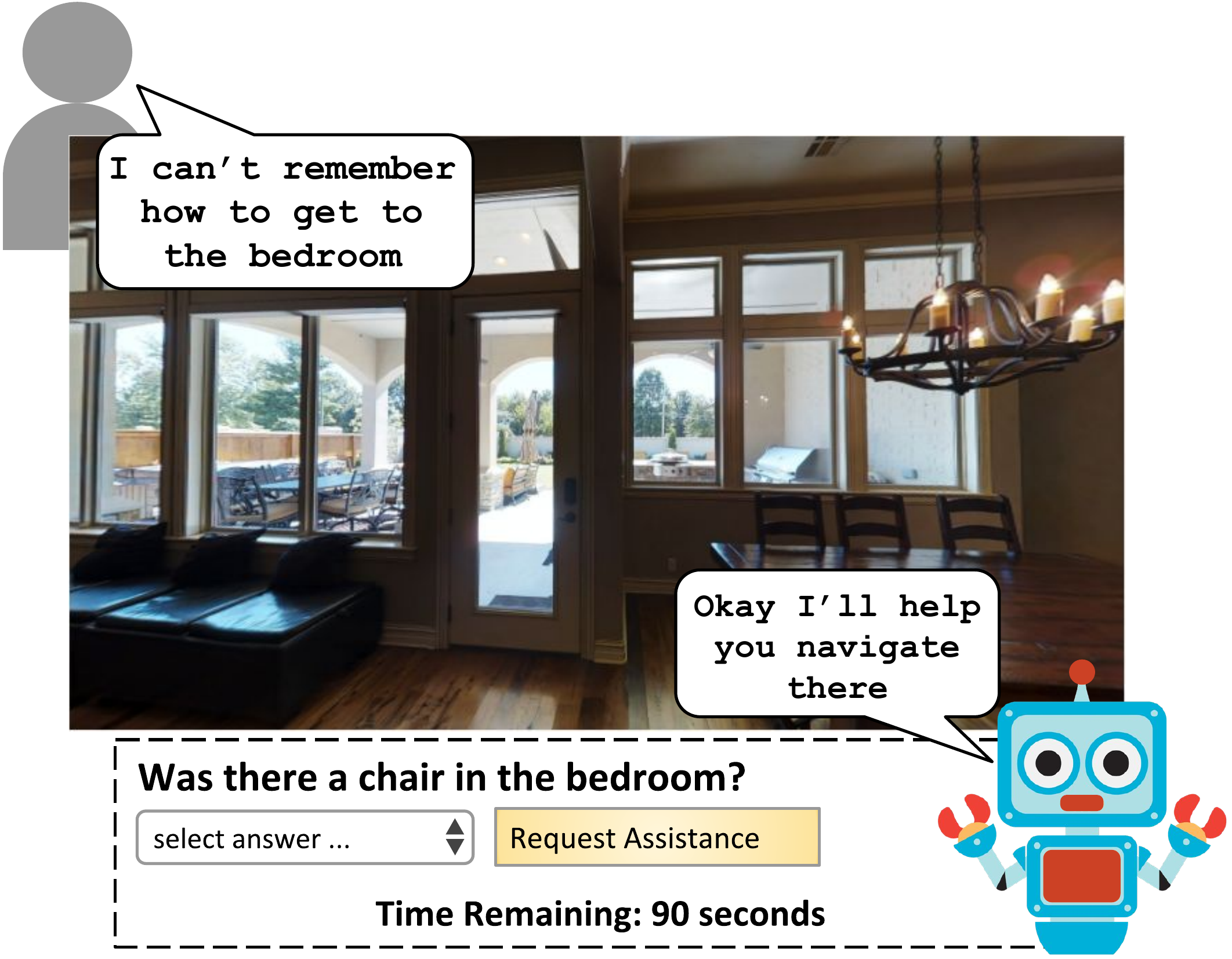}
\caption{Conceptual depiction of the study with 3D environment, option for requesting assistance and human-assistant interaction.}
\label{fig:TeaserFig}
\end{figure}

To gain insight into the kind of local assistance people would like to receive, we gave participants exposure to a 3D environment (a "fly-through" of a Matterport3D environment \cite{chang2017matterport3d}). They were then asked questions about the environment, as illustrated in  Figure~\ref{fig:TeaserFig}. For each question, participants could either (1) answer the question immediately, (2) navigate back to the location where the answer could be discerned, or (3) pay for an assistant to bring them back to that location. 

In this paper, we present summary statistics and results of models (which employ hand-selected features) that predict whether participants will ask for assistance. The features used in these models were selected based primarily on intuition. Ultimately, we aim to formulate models of human perceptual and memory systems that are based on established computational models of human perception and cognition (e.g.,~\cite{Chater:2006ku,Knill:1996}). We hypothesize that this knowledge will help address a core challenge of developing accurate priors for when people will ask for assistance in memory and navigation tasks. Such priors provide a foundation for more generalizable models and inferring model parameters from less data (i.e., low-shot learning). 

\noindent In summary, our main contributions are as follows:
\begin{enumerate}[leftmargin=*]
  \item We introduce the Memory Question Answering (MemQA) task for humans, which tests human visual spatial memory. We created the Visually Grounded Memory Assistant Dataset, which contains over 6k instances of humans preforming the MemQA task. To the best of our knowledge, this is the largest dataset on visually-grounded memory assistance for humans.
  \item We perform in-depth analysis of conditions under which humans ask for assistance. 
  \item We develop baseline models for the task of predicting whether participants will ask for assistance or navigate on their own as well as their accuracy on answering the MemQA questions. 
\end{enumerate}

\section{Related Work}
\subsection{Research and models of human memory}
Human memory is often classified based on the length of storage (sensory, short, and long term) and the ability to communicate the contents of memory. While contents of \textit{declarative} memory can be stated explicitly, \textit{procedural} or perceptual-motor skills cannot be so stated~\cite{Baddeley:1999tm}. \textit{Declarative} memory is often further classified into \textit{episodic} (life events) and \textit{semantic} (language and symbol-based knowledge). In addition to this classification scheme, computational theory has led to the development of process models for the encoding, storage, and retrieval of memories. These models, typically built on the basis of association networks, blur the lines in the typology and capture important patterns in data~\cite{Howard:2017du,Howard:2014iq}. In particular, the encoding and recall of memories is highly dependent on spatio-temporal context and the memory networks built with this structure seem to associate representations across the memory typology described above. For example, people can often report the context in which they learned to tie their shoe (procedural and episodic memory) and shoe brands are likely faster to recall when a person is tying their shoe than when they zipping up their jacket (indicating context specific effects of procedural and semantic memory). 

The complexity of these association networks has, historically, been difficult to study in detail due to methodological limitations. In particular, studying memory at scale and in real-world, visually-rich contexts is a significant challenge. By collecting data on visual memory tasks at scale on Amazon Mechanical Turk (AMT) in complex 3D environments, our data set provide an important step toward overcoming these limitations. AMT studies have also been used in the past to study memorability of images~\cite{goetschalckx2019ganalyze, khosla2015understanding}. However, these studies focus on purely visual features underlying memorability and do not account for context-driven or task-driven visual memory encoding.

The most relevant cognitive-science research for our task focuses on how people learn to navigate environments. People build and store mental maps that allow for more efficient navigation on future visits to that location. These maps are built from mixtures of sensory cues that include landmarks, optical flow as well as non-visual cues~\cite{Etienne:2004fg,Zhao_vx,Gillner:1998dd}. What features are used and how they are encoded is not fully understood~\cite{Collett:2013cr,Jetzschke:2017gk,Chan_kd}. Our data provide a rich source of information to develop our understanding of what visual features are encoded and stored in human spatial maps. Understanding what features are used by humans may enable the development of better navigation systems in mechanical autonomous agents. Such agents are now being developed in a research program on embodied question answering (EQA)~\cite{embodiedqa, yu2019cvpr, wijmans2019embodied} which was the primary machine-learning inspiration for the present study.

\begin{comment}
\todo{Given Kevin's response to the idea that navigation was relevant, do we need to say something like "Despite the fact that people do not choose a navigation path during the fly-through in our study, research on how people learn to navigate 3D environments is highly relevant to the present study."}
\end{comment}

\subsection{Embodied Perception and Question Answering}
Recently, the computer-perception community has opened a new field of embodied perception where agents learn to perform tasks in 3D simulated environments in an end-to-end manner from raw pixel data. These tasks include target-driven navigation~\cite{zhu2017target}, instruction-based visual navigation~\cite{anderson2018vision}, and embodied and interactive question answering (EQA)~\cite{embodiedqa, yu2019cvpr, wijmans2019embodied}. In a typical EQA task set up, an agent is spawned in a random location of a novel building and asked a question about an object or room such as ``What color is the car?''. The agent has no prior knowledge or representation of the building or objects and it must navigate to find the object and then answer the question correctly. This task involves learning a robust navigational system and an accurate visual inference to answer the question. This task was designed as a good measure of an agent's ability to preform visually grounded navigation and semantic understanding of the environment. We hypothesize that observing humans in the EQA task will lend insight into human spatial memory and semantic understanding of the environment. To this end, we expanded the EQA dataset to include five types of questions: location, existence, color, count, and comparison. We then use the new EQA questions to create a new task for humans called Memory Question Answering (MemQA). In the MemQA task, participants are given a short video of a fly through of an environment and are then asked to solve multiple EQA questions about that environment within a time constraint.

Apart from enabling the study of how humans perform navigation tasks and encode spatio-temporal information, our task also allows humans to ask for assistance when needed. Recent research in embodied and autonomous agents has also explored the utility of seeking assistance~\cite{Nguyen:2019vp, Nguyen:2019wa}. Nguyen and Daumé III developed a navigation task where agents could ask for natural language assistance~\cite{Nguyen:2019vp}. Their goal was to develop mobile agents that can leverage help from humans potentially to accomplish more complex tasks than the agents could entirely on their own. They also explored using a cost to for each request for assistance. Their goal was to try and learn the optimal policy for requesting assistance with a limited budget of requests. They did not gather human assistance dialogue but instead supplemented the assistance with the instructions from the Room2Room task~\cite{anderson2018vision}. Unlike this work, our focus is on understanding when humans might seek assistance in tasks that require visual memory encoding. The ability to understand when a user has forgotten something about the local environment and thereby might need assistance would be crucial for the next generation of contextual, personal assistants.

\subsection{Automated interaction and AI Assistance}
Research in human--robot interaction and simulations of human--AI cooperative systems has helped identify and make progress toward some of the major challenges in automated interaction systems. As noted above, our data provide a foundation for models that better predict what information people encode and remember during navigation. Under the assumption that humans use informative features to learn spatial maps, identifying and using these features to train autonomous AI agents is likely to lead to more efficient learning than learning from raw pixels. Some research toward this goal has used simulated human agents to answer questions from an autonomous agent to help it learn to navigate~\cite{Nguyen:2019wa,Nguyen:2019vp}. A limitation of using simulated human agents is that AI--AI learning does not necessarily translate into improved human--AI performance~\cite{Chattopadhyay:2017ws}, a fact that has been starkly revealed in the development of autonomous vehicles~\cite{Sadigh:2016um,Sadigh:2017cz}. This brings us to the second applications of research from our data set: The development of effective AI assistance using first-person camera data.

By understanding perceptual features that humans use to perform a task (i.e., a `theory of mind' for human perception and memory in our case), an automated assistant will be able to better identify when and how to intervene with assistance. Recent studies where the AI agent has third-party observation have demonstrated that this `theory of mind' approach with the assumption that humans will act rationally to achieve a goal, has advantages for the development of human--AI collaboration~\cite{Liu:2016do,Baker:2017is}.  When the agent has first-person video, this approach lays the foundation for understanding what features of the environment and what task a person is likely to attend to (see~\cite{Betancourt:ct6yR1_t,Nguyen:2016ee} for recent reviews of computer-vision approaches to action understanding from first-person video). A machine-based representation space that is better aligned with human perceptual and memory representations allows for better grounded interaction and communication between the human and AI agents. We hypothesize that identifying this common representation space will lead to better generalization of models for visually grounded assistance. We further envision that our data would provide a foundation for testing current models of human perception and memory from cognitive science. If such models are predictive, they can and should be incorporated into a `human-like' representation for visually-grounded AI assistants. 

\section{The Visual Assistance Dataset}\
We now describe the Memory Question Answering task and the data set collection protocol for the Visually Grounded Assistant (VGA) Dataset. To re-iterate, the goals of our dataset collection were (1) to provide a basis for the development of an agent that can predict when a person is likely to request assistance and (2) to develop and test models of human visual and spatial memory.

\subsection{Task Description}
\label{sec:task-description}

\textbf{Task for Human} We refer to the task for the human participants as Memory Question Answering with Navigation (MemQA). Please refer the video\footnote{\url{https://www.youtube.com/watch?v=T97r2leqFyQ}} to see a demonstration of the MemQA task, where the assistant is an oracle that can be used to help answer questions about the environment. The task was encapsulated in the 3D simulated indoor environments from the Matterport3D dataset~\cite{chang2017matterport3d} and was run on Amazon Mechanical Turk. The Matterport3D dataset has been used in many embodied perception tasks~\cite{wijmans2019embodied,anderson2018vision,nguyen2019cvpr} and thus our data from humans complements existing data from AI agents. Additionally we used the Matterport3D dataset because it contains scans of real buildings with realistic settings which range from offices to houses. The setup of our data collection interface is a simulation which uses the actual RBG panoramic frames of the environment. This removes any negative impact that reconstruction errors or unrealistic looking scenes could have on a participants memory. 

\begin{figure*}[htp]
\centering
\includegraphics[width=\textwidth]{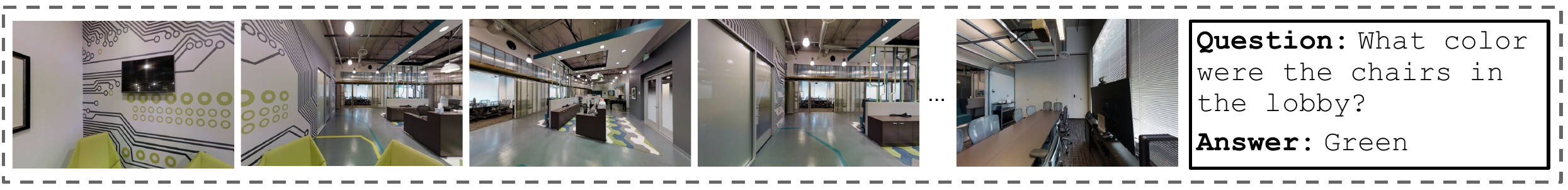}
\caption{Frames from a fly through and visual question from the MemQA task. These panoramic RGB images of a Matterport3D building were rendered using the Matterport Simulator.}
\label{fig:flythrough}
\end{figure*}

In each trial, the participant is exposed to a fly through of an indoor 3D simulated building as depicted in Figure \ref{fig:flythrough}. This fly through is on average a little longer than 70 seconds. After the fly through, the participant is teleported back to the starting point and is presented with four questions. The questions are always about objects and rooms which were passed during the fly through. The questions can concern the location, color, count, and existence of objects, as well as room comparisons and object comparisons (see Section~\ref{sec:data-collection} and Table \ref{tab:question-types} for more examples and details about the questions). Participants are allotted a total of 2.5 minutes to answer all four questions. They can take three distinct approaches to answer a question: (1) answer immediately without any form of navigation, (2) navigate through the environment themselves (presumably to return to the location where the answer could be discerned), or (3) request assistance, in which case an assistant would transport them to the location from which the answer could be discerned. Selecting the third option carried a time cost of 10(+) seconds, which was subtracted from the time limit. The exact cost of each request for assistance was a function of total distance the assistant would have to navigate them. When the participant requested assistance they would specify which question they wanted assistance with and if it was a multi location question such as object comparison they could select which object they wanted to be navigated to. For example for the question ``were the stove and the bathtub the same color?'' the participant would have to choose whether they wanted to be assisted with finding the bathtub or the stove. In order to incentivise the importance of answering correctly, the participant is given a monetary bonus for each correct answer. \\

\textbf{Task for Assistant} The main goal of the assistant is to work collaboratively with the human to answer the questions. The agent needs to be able to understand the humans behavior so that it can realize when the human needs and wants help while not being overly intrusive. To do this the agent needs to model an an accurate representation of what a human has encoded during the fly through and to predict the humans behavior during the question answering phase. The assistant agent is set to act as an oracle for the building and is allowed access to the annotated 3D mesh as well as the questions the human is answering. This dataset creates multiple interesting tasks for an assistant agent. The tasks we propose are most useful to realization of an AR assistant are:
\begin{enumerate}

\item Take in the fly-through and a single question. Predict: correctness, navigation behavior, assistance request behavior.
\item Take in the fly-through, all four questions and the sequence of frames during the answering phase. At each time step of the answering phase predict behavior: navigation, request for assistance, answer selection or nothing
\end{enumerate}
In Section 5 we define a method for the first task and give some insight of methodologies and metrics for solving and evaluating the second task.

\subsection{Dataset Collection}
\label{sec:data-collection}
\noindent\textbf{Simulators}\\
The Matterport3D dataset \cite{chang2017matterport3d} includes over 10k panoramic RGB-D images over 90 real indoor buildings. These panoramic nodes are distributed on average 2.25 meters apart across the entire building. The Matterport simulator~\cite{anderson2018vision} creates an interactive and navigable environment for the Matterport3D dataset. We used the Habitat simulator~\cite{habitat} to extract additional information, such as semantic segmentation of the mesh.\\

\noindent\textbf{Question and Fly through Creation}\\
Following the methodology from \cite{embodiedqa}, we generated the questions programmatically using the Matterport3D meshes and annotations. Each question was represented as a functional template as shown in Table~\ref{tab:question-types}. Each template defines the query-able rooms or objects. The original EQA dataset on Matterport3D contained the question types indicated by the $*$ superscript. To obtain a more diverse and representative question set, we added existence, count, and comparison questions. These questions provide additional data to determine how difficult different features of the environment are for people to encode and remember. In order to ensure consistency of answers, participants selected the answer from a drop down menu of all possible answers. The random chance accuracy for all questions in the dataset is 29.08\%.

After the initial question generation, we discovered many errors in the Matterport3D annotations. To eliminate erroneous annotations, we ran a crowd-sourced study to verify annotation accuracy. Table~\ref{tab:question-types}, lists the number of original generated questions, the number that were filtered by the verification study and the remaining number of questions. This study resulted in the necessary filtering of ~20\% of the generated questions. 

%Flythrough path generation
We generated the fly-through paths to ensure that they included the visual information needed to answer the questions with a minimum-distance criteria. The fly-through paths were generated after filtering and generating the questions. To generate each fly through, we first randomly sampled, without replacement, 4 questions about different objects from the same environment. We then computed the shortest path through the environment that visited each required location. A short random trajectory was added to the start and end of the path. For purposes of consistency, fly throughs under 45 seconds or over 75 seconds were discarded.

\begin{table*}
  \caption{MemQA question types and templates. The $^*$superscript denotes question types that were included in the original Matterport EQA~\cite{wijmans2019embodied}.}
  \label{tab:question-types}
  \begin{tabular}{ccccl}
  \toprule
    \textbf{Question Type} & \textbf{\# Generated} & \textbf{\# After Filtering} & \textbf{Template}\\
    \midrule
\texttt{$^*$location} & 203 & 116 &\textit{What room is the <OBJ> located in?}\\
\texttt{$^*$color} & 299 &  188 &\textit{What color is the <OBJ>?}\\
\texttt{$^*$color inroom} & 1432 &  943 &\textit{What color is the <OBJ> in the <ROOM>?}\\
\texttt{existence} & 283 &  207 &\textit{Is there a <OBJ> in the <ROOM>?}\\
\texttt{count object} & 2729 &  2463 & \textit{How many <OBJS> in the <ROOM>}?\\
\texttt{count room} & 340 &  299 & \textit{How many <ROOMS> in the house?}\\
\texttt{color compare inroom}  & 320 & 288 &\textit{Does <OBJ1> share same color as <OBJ2> in <ROOM>?}\\
\texttt{color compare xroom}  & 86 & 70 &\textit{Does <OBJ1> in <ROOM1> share same color as <OBJ2> in <ROOM2>?}\\
\texttt{object size inroom}  & 300 & 272 &\textit{Is <OBJ1> bigger/smaller than <OBJ2> in <ROOM>?}\\
\texttt{object size xroom}  & 260 & 212 &\textit{Is <OBJ1> in <ROOM1> bigger/smaller than <OBJ2> in <ROOM2>?}\\
\texttt{room size compare}  & 1048 & 932 &\textit{Is <ROOM1> bigger/smaller than <ROOM2> in the house?}\\
    \bottomrule
  \end{tabular}
\end{table*}

\begin{figure}
\centering
\begin{subfigure}{\linewidth}
\includegraphics[width=8.5cm]{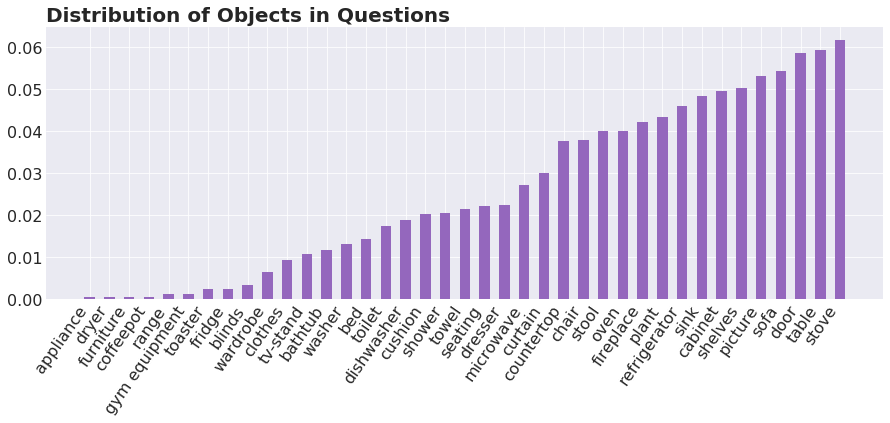}
\end{subfigure}%

\begin{subfigure}{\linewidth}
\includegraphics[width=8.5cm]{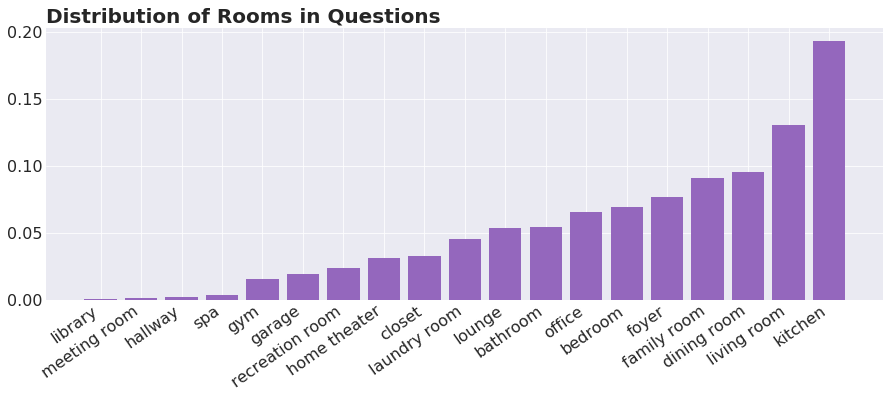}
\end{subfigure}%
\caption{Distribution of the question topics: Each question type is generated from a template and query-able objects and rooms. The graphs show the distribution of query-able objects and rooms we used to generate questions. These distributions of questions were obtained after filtering out the questions that included erroneous annotations in the Matterport3D data set.}
\label{fig:roomDist}
\end{figure}

\section{Visual Assistance Dataset Analysis}

\noindent\textbf{General Statistics}\\
We collected $\geq 5$ annotations for 1250 unique MemQA tasks with a total of 6275 annotations. Note that each MemQA task consists of a single fly through and four questions as described in Section~\ref{sec:task-description}. Table~\ref{tab:general-stats} shows some basic statistics of the task and data collected. 

\begin{table}
  \caption{VGA Dataset: General Statistics}
  \label{tab:general-stats}
  \begin{tabular}{ll}
    \toprule
    Avg. length of fly through (seconds)& 71.45s\\
    Avg. time taken on task (seconds)& 74.96s\\
    Percent of annotations that reached the time limit& 9.68\%\\
    Num. of unique workers& 412\\
    Avg. participant accuracy & 70.34\% \\
    Random Chance accuracy & 29.08\%\\
    Percent of annotations that used assistance& 58.34\%\\
    Percent of questions with assistance requests& 25.23\%\\
  \bottomrule
\end{tabular}
\end{table}

While the average time taken on the task was 74.96 seconds, on 9.68\% of annotations a participant reached the time limit. This almost always occurred on annotations where participants requested assistance. This shows that the time limit acts as an effective way of budgeting the total number of assistance requests thus disallowing participants from completely relying on the assistant.

\begin{figure}
\centering
\includegraphics[width=8cm]{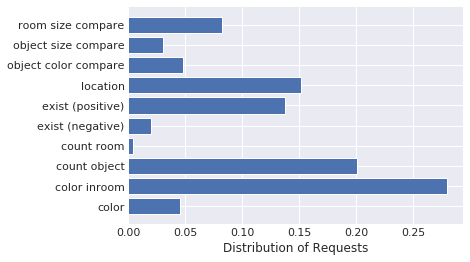}
\caption{Distribution of Assistance Requests by Question Type: The frequency of requests for assistance is a good indicator of question difficulty.}
\label{fig:assistReqPerQType}
\end{figure}

\begin{figure}
\centering
\includegraphics[width=8.5cm]{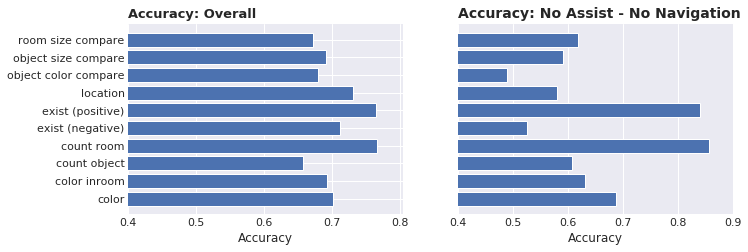}
\caption{Accuracy by Question Type: The left panel shows the accuracy for each question type for the entire data set. The right panel shows accuracy for questions that were answered without assistance or navigation.}
\label{fig:accByQType}
\end{figure}

The differences in question difficulty is coarsely exposed through Figure~\ref{fig:assistReqPerQType} that shows the proportion of requests for assistance and Figure~\ref{fig:accByQType} that shows accuracy by question type. The fewest number of requests came for room count and nonexistence questions. While this suggests that these questions were the easiest to answer, the latter fact likely reflects the fact that people recognized that they would not gain any information by being brought to the absence of an object. Comparison questions also had few requests. This could reflect either difficulty level \textit{or} a lack of willingness to spend the time required to go to multiple locations to answer the question. This is where our dataset provides opportunity to examine trade-offs between the time/effort it would take to find out the right answer and risking a guess to save that time. The most assistance was requested for room in color and object count suggesting these were difficult questions. Similar trends are observed in the accuracy data, which we examine next.  

The left panel of Figure \ref{fig:accByQType} shows accuracy of all answers (with or without assistance) and the right panel shows accuracy only for questions when the participant neither asked for assistance nor navigated to the location to obtain the answer.  The right panel reveals clearly that room count and object existence questions were the easiest for people to remember. Interestingly, there is a substantial difference between their ability to answer existence and non-existence questions. This may reflect participants' recognition that there is a good chance that they missed something that was present in the environment, making it more likely that they will guess it exists when, in fact, it doesn't exist.

\begin{figure}
\centering
\begin{subfigure}{\linewidth}
\centering \includegraphics[width=7cm]{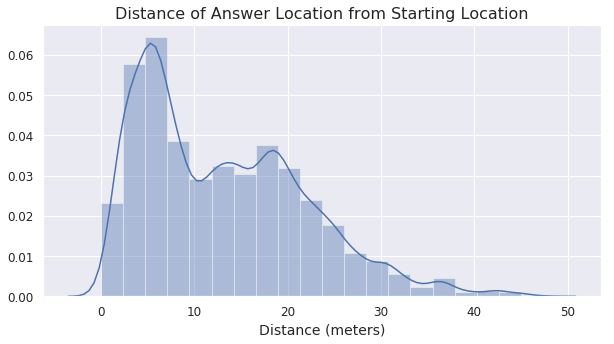}
\end{subfigure}
\begin{subfigure}{\linewidth}
\centering \includegraphics[width=7cm]{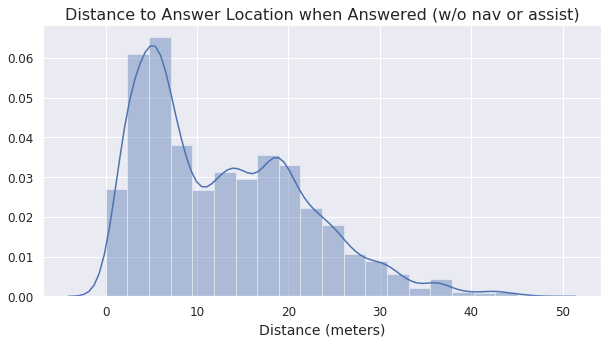}
\end{subfigure}
\begin{subfigure}{\linewidth}
\centering \includegraphics[width=7cm]{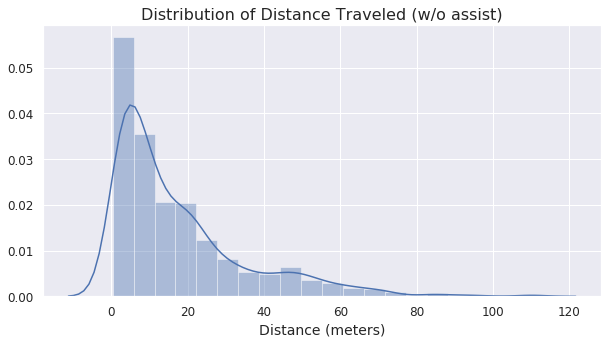}
\end{subfigure}
\begin{subfigure}{\linewidth}
\centering \includegraphics[width=7cm]{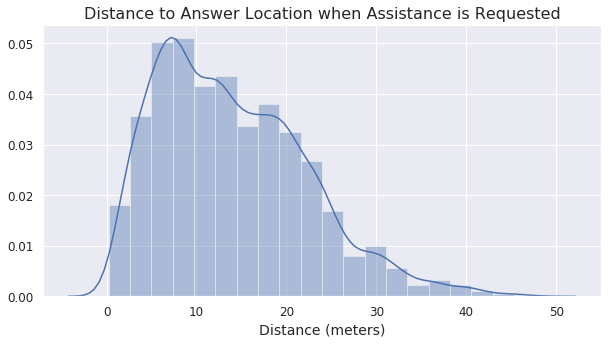}
\end{subfigure}
\caption{Distribution of Distance to Target under different condition: The distance in meters from the participant to the location answer. Illustrates the location conditions under which participants ask for assistance.}
\label{fig:assistDist}
\end{figure}

Finally, we wanted to see if the distance to the target location effected choices to navigate or get assistance. Figure \ref{fig:assistDist} shows the distribution of distances to all targets (1) from the starting point, (2) for questions that were answered without navigation or assistance, (3) for cases where participants navigated to the target and (4) for cases where assistance was requested. While there are long tails in each of these distributions, the cases where people navigated to the target is the only distribution with a clear mode at the shortest distance. This suggests that participants were reluctant to navigate to targets when travel distance was long. To further examine the effect of distance, we included `time of first exposure', which is correlated with distance, as a feature in the models we present in the next section. 

\section{Baselines and Methods}
We explore modeling the first task described in Section \ref{sec:task-description}: given the fly through and question, predict whether a participant will be able to answer the question correctly, and whether they will answer the question (1) without any form of navigation, (2) by navigating on their own, or (3) by requesting assistance. We adopt a modeling approach that consists of constructing four binary classification problems: one for answer correctness, and one corresponding to each of the three answer-strategy approaches mentioned above. While we could construct a single 3-class classifier to predict the participant's strategy, a collection of binary classifiers enables more nuanced study of which outcomes are easiest to predict; for example, we can generate a receiver operating characteristic (ROC) curve for the classifier associated with the prediction of each outcome.

The goal of this study is to model human visual and spatial memory. The participant is always exposed to the correct answer of a question during the fly through. Whether the participant is able to answer the question after completing the fly through is a direct measure of their ability to encode and recall the relevant feature of the environment. Additionally, the participant's ability to navigate back to the location of the answer is informed by their spatial memory. Visually encoding all of the information into memory is outside the bounds of human memory ability. What features actually get encoded will depend on many factors such as object saliency and the duration of the fly through. By modeling the participants' performance based on the fly through alone, we seek to determine the most important factors in determining performance. For the initial model, we did not consider question types that dealt with multiple objects. This is, we omitted comparison and count questions, resulting in five remaining question types. 

Let us call primary object of the question $o_1$. We started by characterizing each question with the following (hand-selected) features:
\begin{enumerate}
\item\label{feat:qType} Type of question (discrete).
\item Length of fly through (continuous).
\item Time of first exposure to $o_1$ (continuous).
\item The temporal exposure of $o_1$.
\item \label{feat:spatExposure}The spatial exposure of  $o_1$ (continuous).
\item\label{objectType} Word embedding of object type of $o_1$ (continuous).
\end{enumerate}

\begin{figure}
\centering
\begin{subfigure}{.5\linewidth}
\centering \includegraphics[width=4cm]{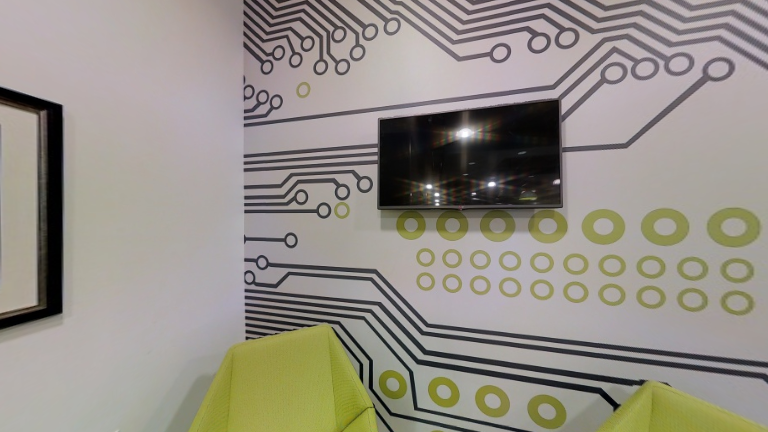}
\caption{Panoramic RGB}
\end{subfigure}%
\begin{subfigure}{.5\linewidth}
\centering \includegraphics[width=4cm]{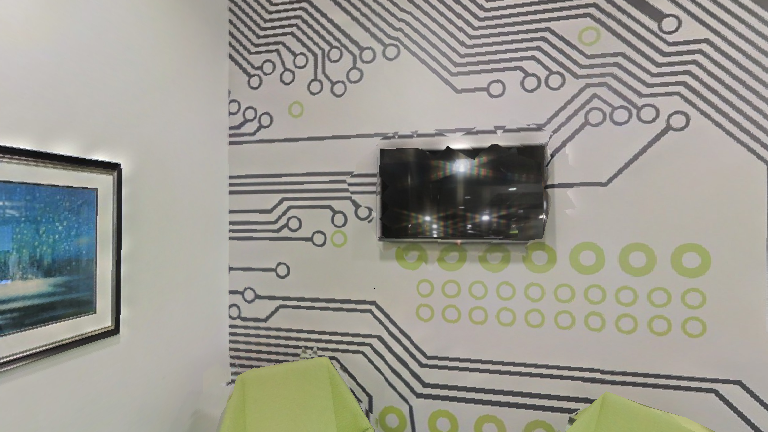} \caption{Reconstruction RGB}
\end{subfigure}
\begin{subfigure}{.5\linewidth}
\centering \includegraphics[width=4cm]{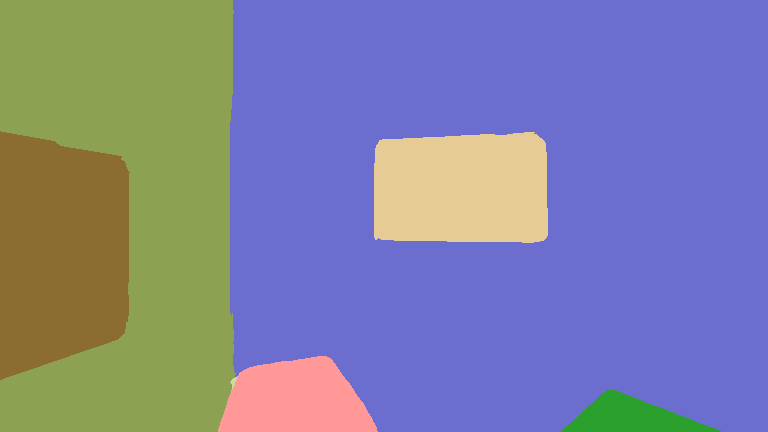} \caption{Instance Segmentation}
\end{subfigure}%
\begin{subfigure}{.5\linewidth}
\centering \includegraphics[width=4cm]{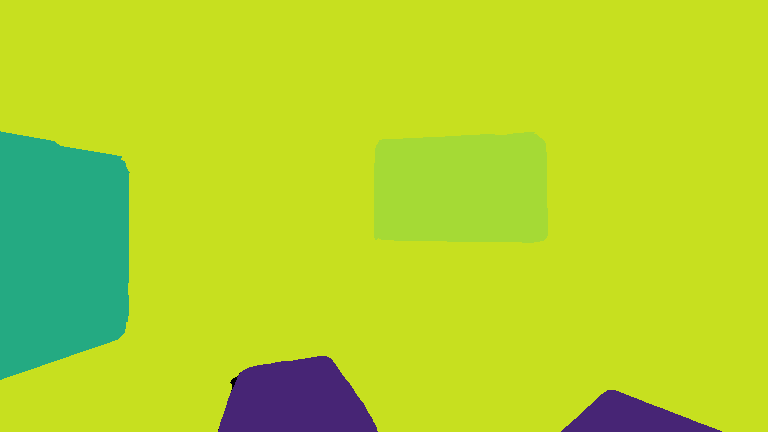}
\caption{Semantic Segmentation}
\end{subfigure}%
\caption{Example of four frames types for the same camera view point. (a) is an example of the view that participants had in the MemQA task. (b) shows the Habitat Simulator reconstruction from the environment mesh. The mesh and it's annotations were used to construct the MemQA questions as well as to create the features for modeling accuracy and assistance requests. The (c) and (d) show segmentation as annotated by the environment mesh.}
\label{fig:examples}
\end{figure}

Features 3, 4, and 5 were drawn from the Habitat simulator over the frames of the fly. Using the instance and semantic segmentation over each frame, the object information from each view was extracted (Figure~\ref{fig:examples}). To obtain a measure of exposure of objects, it was necessary to consider exposure both in temporal and spatial terms. Temporal exposure refers to the length of time the object was in view and spatial exposure refers to the amount of area the object occupied throughout the fly through. The spatial exposure is defined as the total of number of pixels of $o_1$ across all frames of the fly through. 

\label{sec:method}
\subsection{Model Description}

As mentioned above, we used these features to train four binary classifiers. We now describe the features and classification models we consider:

\textit{Features}. We employ 10 features in the models:  the features \ref{feat:qType}--\ref{feat:spatExposure} above, as well as a 5-dimensional embedding of the object type derived from feature \ref{objectType} above; note that the dimension of the embedding space is a hyperparameter, which we set to five to keep the number of features relatively small. To compute this 5-dimensional embedding, we (1) compute the GloVe embedding \cite{pennington2014glove} of all 35 unique object types into a 50-dimensional latent space, (2) apply principal component analysis (PCA) \cite{abdi2010principal} and project the embeddings onto the 5-dimensional linear subspace of this 50-dimensional latent space spanned by the first five principal components, and (3) treat the resulting five PCA coordinates of the object type as features.

\textit{Classification models}. We consider three different classification models as implemented in Scikit-learn: 
\begin{enumerate}
    \item \textbf{Random-forest (RF) classifier}. We employ an ensemble of 10 trees and we use the Gini impurity measure in tree construction; all remaining parameters correspond to the default values in Scikit-learn.
    \item \textbf{Multilayer-perceptron (MLP) classifier} (i.e., feedforward, fully connected neural network). We employ one hidden layer with 100 neurons, ReLU activations, and the Adam optimizer; all remaining parameters correspond to the default values in Scikit-learn.
    \item \textbf{Support-vector-machine (SVM) classifier}. We employ a radial-basis-function kernel with coefficient $\gamma = 2$ and penalty parameter of $C=1$; all remaining parameters correspond to the default values in Scikit-learn.
\end{enumerate}
Models are trained using 80\% of the data, and are tested on the remaining 20\%. We ensure that the training and testing sets contain different questions, such that results on the test set assess generalization of the models across both participants and questions.

\subsection{Results}

\begin{figure}[t!]
    \centering
    \begin{subfigure}[t]{0.23\textwidth}
        \centering
        \includegraphics[width=\textwidth]{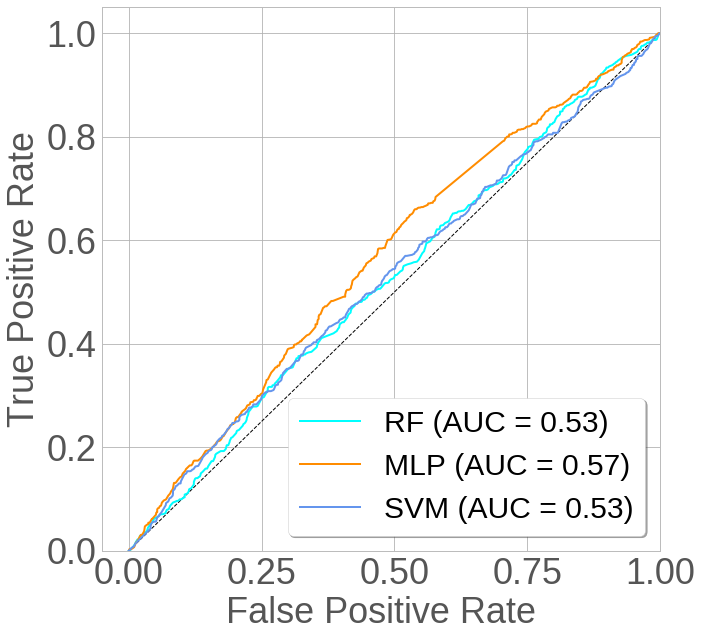}
        \caption{ROC curves for models predicting if the participant answered the question without any form of navigation}
    \end{subfigure}\ 
    \begin{subfigure}[t]{0.23\textwidth}
        \centering
        \includegraphics[width=\textwidth]{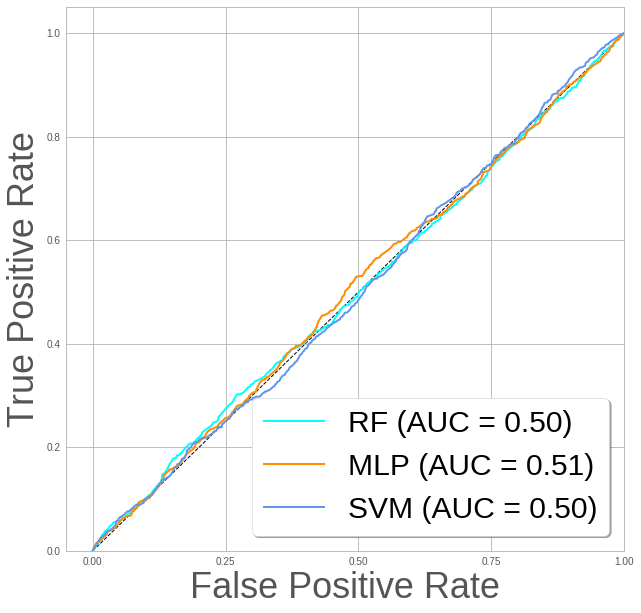}
        \caption{ROC curves for models predicting if the participant answered the question by navigating on their own}
    \end{subfigure}\ 
    \begin{subfigure}[t]{0.23\textwidth}
        \centering
        \includegraphics[width=\textwidth]{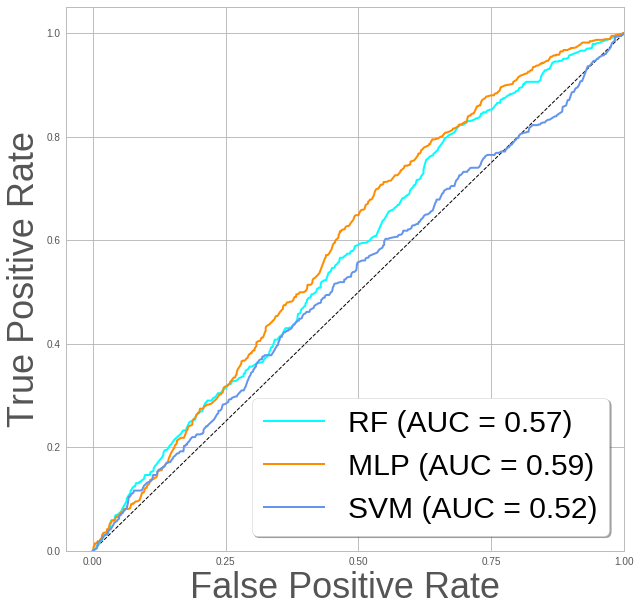}
        \caption{ROC curves for models predicting if the participant answered the question by requesting assistance}
    \end{subfigure}\ 
    \begin{subfigure}[t]{0.23\textwidth}
        \centering
        \includegraphics[width=\textwidth]{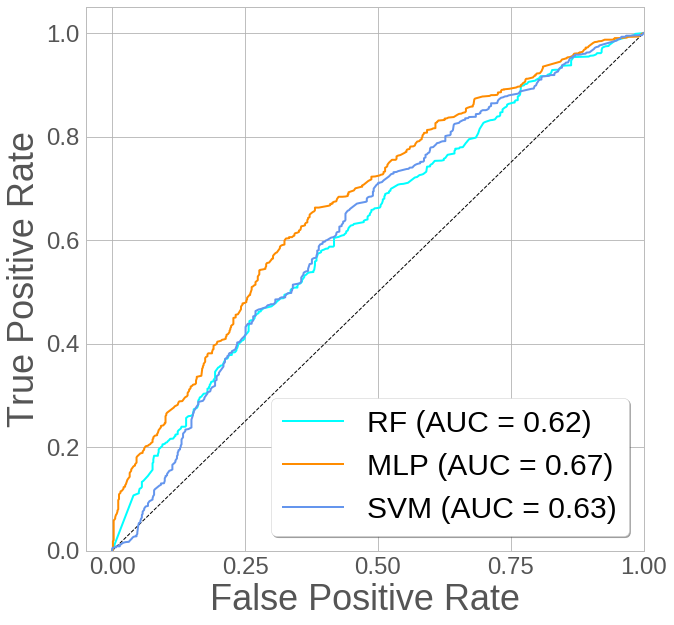}
        \caption{ROC curves for models predicting if the participant answered the question correctly}
    \end{subfigure} 
    \caption{ROC curves for the four types of predictions and three considered classification models}
    \label{fig:AUCROC}
    \end{figure}

Figure \ref{fig:AUCROC} reports the receiver operating characteristic (ROC) curves for each of the candidate classification models on each of the four binary prediction tasks computed on the test set, with Table \ref{tab:AUCROC} reporting the associated area under the curve (AUC).
    
\begin{table}[htp] 
\centering 
\begin{tabular}{l|c|c|c}  
\hline
\textbf{Response} & \textbf{RF} & \textbf{MLP} & \textbf{SVM} \\
\hline
\midrule  
no navigation & 0.53 & 0.57 & 0.53 \\
self navigation & 0.50 & 0.51 & 0.50 \\
assistance & 0.57 & 0.59 & 0.52 \\
answered correctly & 0.62& 0.67 &0.63 \\
\hline  
\end{tabular}  
\caption{AUC--ROC values for the candidate classifiers and prediction tasks (RF: random forest; MLP: multilayer perceptron; SVM: support vector machine)}  
\label{tab:AUCROC}  
\end{table} 
These results demonstrate that---using the ten features specified above---the models are best able to discriminate whether the participant correctly answered the question; the MLP classifier achieved an AUC--ROC value of 0.67 in this case. The models' next best performance occurs for predicting either the participants' ability to answer with no navigation or with assistance; the MLP classifier achieved AUC--ROC values of 0.57 and 0.59, respectively, in these cases. None of the models performed better than chance in predicting whether the participant employed self-guided navigation to answer the question. Note that currently, none of our features capture the spatial mapping of the environment or the complexity of the fly through path, both of which might be important for predicting whether the users would employ self-navigation. Apart from developing more sophisticated models, we also plan to explore human perception and cognition inspired features in the future. For instance, object saliency has been found to be important for recalling the object~\cite{einhauser2008objects}. Likewise, landmarks, optical flow, and other non-visual cues have been found to be important for navigation~\cite{Etienne:2004fg,Zhao_vx,Gillner:1998dd}.

\begin{figure}[t!]
    \centering
    \begin{subfigure}[t]{0.23\textwidth}
        \centering
        \includegraphics[width=\textwidth]{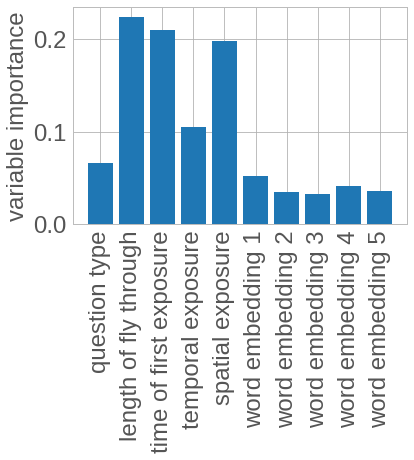}
        \caption{Variable importance for ``assistance''}
    \end{subfigure}\ 
    \begin{subfigure}[t]{0.23\textwidth}
        \centering
        \includegraphics[width=\textwidth]{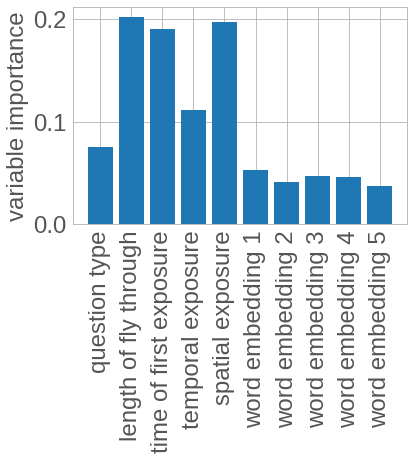}
        \caption{Variable importance for ``answered correctly''}
    \end{subfigure}
    \caption{Variable importance generated by the random-forest classifier for two responses.}
    \label{fig:featureImportance}
    \end{figure}

We now turn to the question of which features are most important for determining the above outcomes. For this purpose, Figure \ref{fig:featureImportance} reports the feature importances arising from the random-forest classifier for responses corresponding to ``assistance''  and ``answered correctly''; we note that the importances for ``no navigation'' and ``self navigation'' are nearly identical to the former. These figures demonstrate that the most important features driving the construction of the decision trees correspond to (1) the total number of time steps in the fly through, (2) the time step at which the object is exposed, and (3) the spatial exposure of the object. Interestingly, the semantic meaning of the objects---as represented by the word embeddings---are characterized by low variable importance; the question type also is relatively unimportant.

\section{Discussion}
The primary purpose of this paper was to introduce a new data set. The data provide an opportunity to develop models of human visually-grounded memory that can serve as a basis for an automated memory assistant. The assistant in our task was an oracle that only responds when called for. However, the simple model we presented above demonstrates promise of developing a model from a richer feature set that can predict when assistance is needed without an explicit request. In addition to the opportunity to develop machine representations that align with human perceptual and memory systems, our data offer an opportunity to examine how people trade off the cost of time to obtain the right answer with the risk of getting the wrong answer (and, hence, not receiving the monetary reward). This could provide a rich avenue for gaining insight into how people value the travel time associated with obtaining the right answer, money, and risk (i.e. the probability of being wrong when guessing an answer without confirmation). Similar situations arise, for example, when people travel to different shops for items at lower prices than the store they are in. 

\section{Conclusion}
The primary aim of this paper was to introduce the `The Visually Grounded Memory Assistant Dataset'. The summary statistics and baseline model demonstrate the potential of using these data to develop models that can predict when people will ask for visual and spatial memory assistance.  

Our dataset creates a rich set of tasks to explore including predicting human performance, behavior and memory. We explored modeling predicting human performance on the MemQA task from what participants were visually exposed to in the fly through as well as the context of the task they were solving. Future work can explore looking at their behavior during the question answering phase and model predictions from raw pixels.

\bibliographystyle{ACM-Reference-Format} 
\bibliography{main}

%%%%%%%%%%%%%%%%%%%%%%%%%%%%%%%%%%%%%%%%%%%%%%%%%%%%%%%%%%%%%%%%%%%%%%%%

\end{document}